\documentclass[10pt,twocolumn,letterpaper]{article}

\usepackage{cvpr}              

\usepackage{graphicx}
\usepackage{amsmath}
\usepackage{amssymb}
\usepackage{multirow}
\usepackage{multicol}
\usepackage{makecell}
\usepackage{booktabs}
\usepackage{algorithm}
\usepackage{algorithmicx}
\usepackage{algpseudocode}
\usepackage[accsupp]{axessibility}

\usepackage[pagebackref,breaklinks,colorlinks]{hyperref}

\usepackage[capitalize]{cleveref}
\crefname{section}{Sec.}{Secs.}
\Crefname{section}{Section}{Sections}
\Crefname{table}{Table}{Tables}
\crefname{table}{Tab.}{Tabs.}

\def\cvprPaperID{7881} 
\def\confName{CVPR}
\def\confYear{2022}

\begin{document}

\title{Fine-tuning Global Model via Data-Free Knowledge Distillation\\ for Non-IID Federated Learning}


\author{{Lin Zhang{$^{1,4}$}}\thanks{Part of this work was done during Ling Zhang's internship at JD Explore Academy.} \quad Li Shen{$^{2}$} \quad Liang Ding{$^{3}$} \quad Dacheng Tao{$^{2,3}$} \quad Ling-Yu Duan{$^{1, 4}$}\thanks{Corresponding author.} \\
$^{1}$\ Peking University, Beijing, China
$^{2}$\ JD Explore Academy, Beijing, China \\
$^{3}$\ The University of Sydney, Sydney, Australia
$^{4}$\ Peng Cheng Laboratory, Shenzhen, China \\
{\tt\small \{zhanglin.imre, lingyu\}@pku.edu.cn, \{mathshenli, dacheng.tao\}@gmail.com} \\
{\tt\small ldin3097@sydney.edu.au}
}

\maketitle

\begin{abstract}
Federated Learning (FL) is an emerging distributed learning paradigm under privacy constraint. Data heterogeneity is one of the main challenges in FL, which results in slow convergence and degraded performance. 
Most existing approaches only tackle the heterogeneity challenge by restricting the local model update in client, ignoring the performance drop caused by direct global model aggregation. 
Instead, we propose a data-free knowledge distillation method to \textbf{f}ine-\textbf{t}une the \textbf{g}lobal model in the server (FedFTG), which relieves the issue of direct model aggregation. 
Concretely, FedFTG explores the input space of local models through a generator, and uses it to transfer the knowledge from local models to the global model. Besides, we propose a hard sample mining scheme to achieve effective knowledge distillation throughout the training.
In addition, we develop customized label sampling and class-level ensemble to derive maximum utilization of knowledge, which implicitly mitigates the distribution discrepancy across clients. Extensive experiments show that our FedFTG significantly outperforms the state-of-the-art (SOTA) FL algorithms and can serve as a strong plugin for enhancing FedAvg, FedProx, FedDyn, and SCAFFOLD. 
\end{abstract}
\section{Introduction}\label{intro}

With the explosive growth of data and the strict privacy-protection policy, reckless data transmission and aggregation gradually become unacceptable due to the high bandwidth cost and risk of privacy leakage. Recently, Federated Learning (FL) \cite{mcmahan2017communication, mills2019communication} has been proposed to replace the traditional heavily centralized learning paradigm and protect data privacy. 
It has been successfully applied in real-world tasks, such as smart city \cite{jiang2020federated,zheng2021applications, qolomany2020particle}, health care \cite{gao2019hhhfl,lu2019learn, liu2021feddg}, and recommender system \cite{hard2018federated,hartmann2019federated}, etc.

One of the main challenges in FL is the data heterogeneity, i.e., the data in clients are non-identically and independently distributed (Non-IID).
It has been verified that the vanilla FL algorithm, FedAvg~\cite{mcmahan2017communication}, leads to drifted local models and forgets the global knowledge catastrophically in this scenario, which further induces degraded performance and slow convergence \cite{hsu2019measuring,lee2021preservation,khaled2020tighter}. This is because the local model is updated merely with local data, i.e.,  minimizing the local empirical loss. However, minimizing the local empirical loss is fundamentally inconsistent with minimizing the global empirical loss \cite{acar2020federated,li2019convergence,malinovskiy2020local} in Non-IID FL.

\begin{figure}[t]
  \centering
   \includegraphics[width=0.95\linewidth]{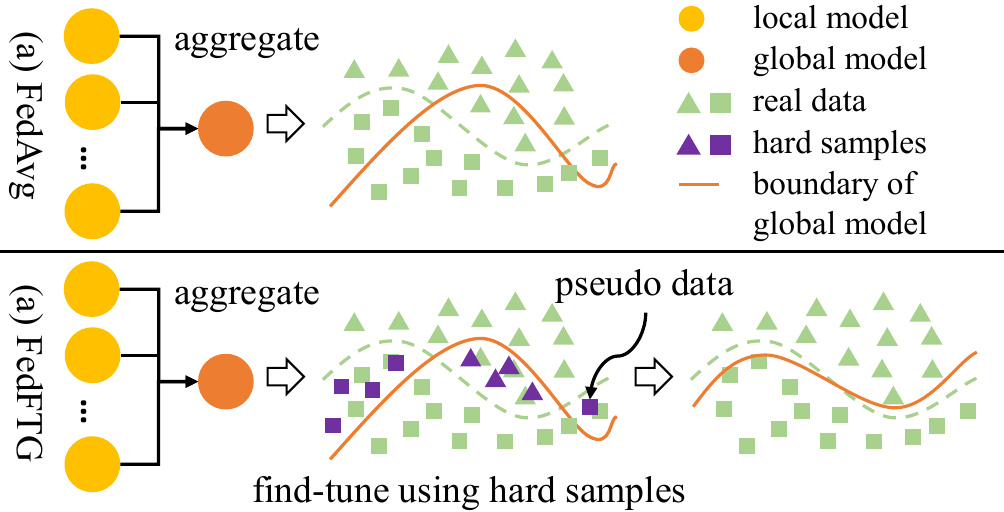}
   \caption{Comparison between FedAvg \cite{mcmahan2017communication} and FedFTG. By fine-tuning the global model using generated hard samples, FedFTG alleviates the performance decrease after model aggregation.}
   \label{fig:sketch}
   \vspace{-10pt}
\end{figure}

To tackle the data heterogeneity challenge, most existing methods, e.g., FedProx \cite{li2018federated}, SCAFFOLD \cite{karimireddy2020scaffold}, FedDyn \cite{acar2020federated}, MOON\cite{li2021model} constrain the direction of local model update to align the local and global optimization objectives. Recently, FedGen \cite{zhu2021data} learns a lightweight generator to generate pseudo feature and broadcasts it to clients to regulate local training. 
However, all these methods merely conduct simple model aggregation to get the global model in server, which ignores local knowledge incompatibility and induces knowledge forgetting in the global model.
In addition, \cite{singh2020model} shows that directly aggregating models will largely degrade the performance while fine-tuning can greatly boost the accuracy. 
These motivate us to fine-tune the aggregated global model in the server with the knowledge in local models. 
On the other hand, merely aggregating local models in server ignores the server's rich computing resources that could be potentially utilized to improve the performance of FL, such as the computing source in cross-silo FL \cite{kairouz2019advances}. 

Motivated by these observations, we propose a novel approach that boosts the performance of standard FL by on-the-fly fine-tuning the global model via data-free knowledge distillation (FedFTG), which simultaneously refines the model aggregation procedure and exploits the rich computing power of the sever. 
Concretely, FedFTG models the input space of local models through an auxiliary generator in the server, then generates pseudo data to transfer the knowledge in local models to the global model to improve the performance. 
To facilitate effective knowledge distillation throughout the training, FedFTG iteratively explores the hard samples in data distribution, which will induce prediction disagreement between local models and global model. 
Figure~\ref{fig:sketch} compares FedFTG with FedAvg. FedFTG fine-tunes the global model with the hard samples to correct the model shift after model aggregation.
The generator and global model are adversarially trained in a data-free manner, thus the whole procedure will not violate the privacy policy in FL. 
Considering the label distribution shift in data heterogeneity scenario, we further propose customized label sampling and class-level ensemble techniques, which explore the distribution correlation of clients and exploit maximum utilization of knowledge.

FedFTG is orthogonal to several existing local optimizers, such as FedAvg, FedProx, FedDyn, SCAFFOLD and MOON, as it only modifies the procedure of global model aggregation in the server. Consequently, FedFTG can be seamlessly embedded into these local FL optimizers, taking their advantages to further improve the performance of FedFTG. Extensive experiments on various settings verify that FedFTG achieves superior performance compared with state-of-the-art (SOTA) methods. 

The main contributions of this work are four-fold:
\vspace{-0.2cm}
\begin{itemize}
\item We propose FedFTG to fine-tune the global model in server via data-free distillation, which simultaneously enhances the model aggregation step and utilizes the computing power of the server. 
\vspace{-0.2cm}
\item We develop hard sample mining to effectively transfer knowledge to global model. Besides, we propose customized label sampling and class-level ensemble to facilitate maximum utilization of knowledge.
\vspace{-0.2cm}
\item We demonstrate that FedFTG is orthogonal to exiting local optimizers and can serve as a strong and versatile plugin to enhance the performance of FedAvg, FedProx, FedDyn, SCAFFOLD and MOON. 
\vspace{-0.2cm}
\item We verify the superiority of FedFTG against several SOTA methods for FL, including  FedAvg, FedProx, FedDyn, SCAFFOLD, MOON, FedGen and FedDF, with extensive experiments on five benchmarks.
\end{itemize}

\section{Related Work}

There exist extensive works on improving the global performance of FL via client selection \cite{fraboni2021clustered,chen2020optimal,huang2020stochastic}, split learning \cite{he2020group,wu2020decentralised}, domain adaptation \cite{peterson2019private,liu2021feddg}, etc. The readers may refer to monographs \cite{kairouz2019advances,wang2021field} and the reference therein to follow up its recent advances. Below, we mainly summarize the most relevant techniques to our work. 

\textbf{Federated Optimizer.} The vanilla FL algorithm, i.e. FedAvg \cite{mcmahan2017communication} periodically aggregates the local models in server and updates the local model with its individual data. 
FedProx \cite{li2018federated} adds a proximal term to the local subproblem to restrict the local update closer to the initial (global) model. SCAFFOLD \cite{karimireddy2020scaffold} uses a variance reduction technique to correct the drifted local update. 
FedDyn \cite{acar2020federated} modifies the objective of client with linear and quadratic penalty terms to align global and local objectives. In summary, all these methods focus on aligning the local and global model to narrow the distribution drift during the local training without enhancing the global model directly as in FedFTG.

 \textbf{Knowledge Distillation in Federated Learning.}
With the help of an unlabeled dataset, FedDF \cite{lin2020ensemble} proposes an ensemble distillation for model fusion, trains the global model using the averaged logits from local models. FedAUX \cite{sattler2021fedaux} finds a model initialization for the local models, and weights the logits from local models using $(\varepsilon, \delta )$-differentially private certainty scoring. FedBE \cite{chen2020fedbe} generates a series of global models from Bayesian perspective using the local models, then summarizes these models into one global model by ensemble knowledge distillation. All these methods rely on an unlabeled auxiliary dataset in the server, while it is unclear to which extent should the auxiliary dataset be related to training data to guarantee effective knowledge distillation. Though FedDF maintains the auxiliary dataset can be replaced with a pretrained generator, it does not instantiate how to acquire the generator. 

\textbf{Data-Free Knowledge Distillation (DFKD).}
DFKD methods \cite{chen2019data,fang2019data} generate pseudo data from a pretrained teacher model, and use them to transfer knowledge of teacher model to another student model. The data is generated by maximizing the response of fake data on teacher model. 
DeepImpression \cite{nayak2019zero} models the output space of teacher model and recovers the real data by fitting the output space. DeepInversion \cite{yin2020dreaming} further optimizes the pseudo data by regularizing the distribution of intermediate feature maps.
DAFL \cite{chen2019data} and DFAD \cite{fang2019data} use a generator to generate data efficiently, where DAFL optimizes the generator by maximizing the response on prediction and feature level, and DFAD uses an adversarial training scheme to exploit the knowledge in teacher model effectively.

FedGen also \cite{zhu2021data} learns a lightweight generator to ensemble knowledge of local models in a data-free manner, but uses the generator to regularize the local training. Besides, we design hard samples mining scheme, customized label sampling, and class-level ensemble to effectively transfer the knowledge from local models to global model in data heterogeneity scenario.

\section{Methodology}

\begin{figure}[t]
  \centering
   \includegraphics[width=0.93\linewidth]{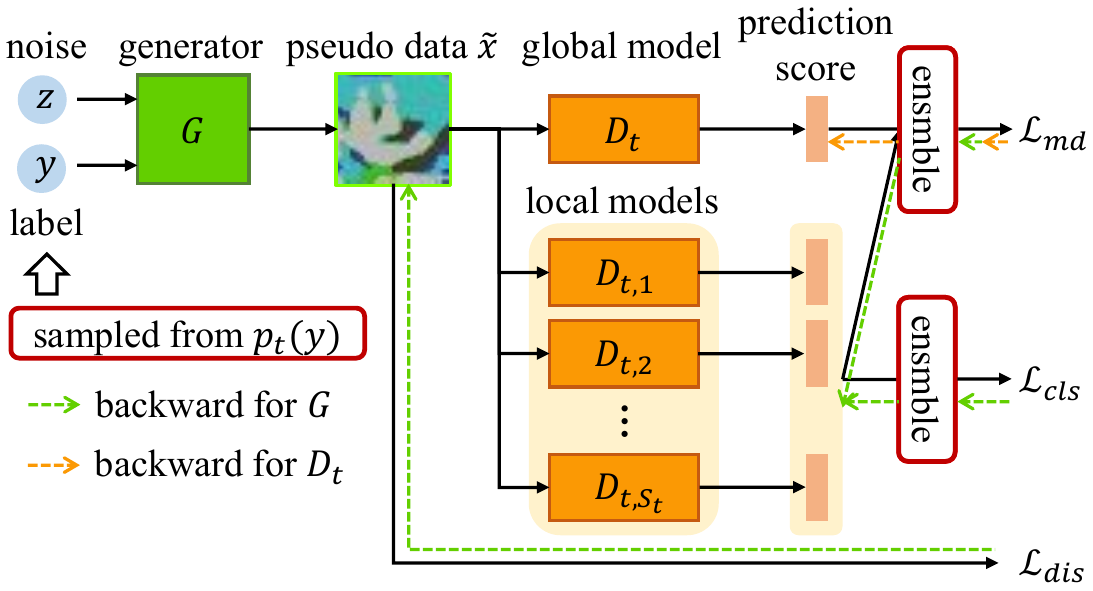}
   \vspace{-0.2cm}
   \caption{Training procedure of FedFTG in the server. After receiving local models in $t$-th round and aggregating them, FedFTG adversarially generates hard sample and transfers knowledge to the aggregated global model through $\mathcal{L}_{md}$. $\mathcal{L}_{cls}$ and $\mathcal{L}_{dis}$ are utilized to boost the fidelity and diversity of hard sample. Besides, FedFTG uses customized label sampling and class-level ensemble to derive maximum utilization of knowledge.}
   \vspace{-7pt}
   \label{fig:overview}
\end{figure}

In this section, we describe the proposed novel federated learning method: FedFTG. 
In each communication round, FedFTG randomly selects a set of clients and broadcasts the global model to them. Each client initializes the local model using the global model and trains it with a local optimizer. The server collects the local models and aggregates them as a preliminary global model. 
Instead of broadcasting the aggregated model back to each client directly, FedFTG fine-tunes this preliminary global model in server using the knowledge extracted from local models.
Concretely, we develop a data-free knowledge distillation method with hard sample mining to effectively explore and transfer the knowledge to global model. Considering the label distribution shift in clients, we propose customized label sampling and class-level ensemble to facilitate more effective knowledge utilization. Figure~\ref{fig:overview} visualizes the training procedure on the server, and the corresponding algorithm is summarized in Algorithms~\ref{alg:flboost}\&\ref{alg:fl-server}.
Note that FedFTG is orthogonal to efforts on optimizing local model training, such as SCAFFOLD, FedAvg, FedProx, and FedDyn.

\subsection{Data-Free Knowledge Distillation With Hard Sample Mining for Global Model Fine-Tuning}\label{sec:3.1}

Let $\omega$ be the model parameter in the server and clients. In this work, we consider there exist $K$ clients, where $\mathcal{D}_k=\{(x_{k,i}, y_{k,i})\}_{i=1}^{N_k}$ is the dataset individually stored in $k$-th client, $N_k$ is the corresponding number of samples. Generally speaking, federated learning can be formulated as the following problem:
\begin{equation}\label{eq:fl_target}
\small
    \min_{\omega} \frac{1}{K}\sum_{k=1}^{K}f_{k}(\omega),\ f_{k}(\omega) = \frac{1}{N_{k}}\sum_{i=1}^{N_{k}}\mathcal{L}(x_{k}^{i}, y_{k}^{i};\omega),
\end{equation}
where $\mathcal{L}$ is the loss function to measure training error, and dataset $\mathcal{D}_k$ for each $k \in \{1,2,...,K\}$ could be distributed heterogeneously. Due to the privacy protection constraint in FL, the server can not directly access local data of clients. To solve Eq.~\eqref{eq:fl_target}, for each communication round $t$, existing methods send the global model $\omega$ to a random set of clients $S_t$ and optimize it by $\min_{\omega}f_{k}(\omega), k \in S_t$. The server collects the local models $\{\omega_k\}_{k \in S_t}$ and aggregates them by averaging the gradients to update the global model $\omega$.

\begin{figure}[t]
\begin{algorithm}[H]
\small
  \caption{FedFTG} 
  \begin{algorithmic}[1]
    \Require
      $T$: communication round; 
      $K$: client number;
      $C$: the fraction of active client in each round;
      $\{\mathcal{D}_{k}\}_{k\in \{1,...,K\}}$: the datasets of clients;
      $\omega$: the parameter of the classifier;
      $\theta$: the parameter of the generator.
    \State initialize model parameters $\omega$ and $\theta$
      \For{$t=1, ..., T$}
        \State $S_t\leftarrow$ (random set of $\left \lceil C\cdot K \right \rceil$ clients);
        \For{$k \in S_t$ \textbf{in parallel}}
          \State $\omega_k\leftarrow$ ClientUpdate$(\omega, D_{k})$ 
          \Statex \Comment{FedAvg, FedProx, FedDyn, and SCAFFOLD}
        \EndFor
        \State $\omega, \theta\leftarrow$ ServerUpdate$(\omega, \theta, \{\omega_k\}_{k\in S_t})$
      \EndFor
  \end{algorithmic}
\label{alg:flboost}
\end{algorithm}
\vspace{-0.4cm}
\end{figure}

\begin{figure}[t]
\vspace{-0.3cm}
\begin{algorithm}[H]
\small
  \caption{ServerUpdate, round $t$} 
  \begin{algorithmic}[1]
    \Require
      $I$: iteration of the training procedure in server;
      $I_g$, $I_d$: inner iteration of training the generator and the global model;
      $\eta_g$: the global step-size;
      $\omega$, $\theta$, $\{\omega_k\}_{k\in S_t}$, $\lambda_{cls}$, $\lambda_{r}$.
    \State $\Delta \omega=\frac{1}{|S_t|}\sum_{k\in S_t}(\omega_k-\omega)$, $\omega\leftarrow\omega+\eta_g\Delta\omega$
    \State compute $p_t(y)$ according to Eq.~(\ref{eq:focusedkd})
        \For{$i=1, ..., I$}
          \State $(Z, Y)\!\leftarrow$ (sample a batch of $z\!\sim\!\! \mathcal{N}(\mathbf{0},\mathbf{1})$ and $y\!\sim\! p_t(y)$)
          \State compute $\{\alpha_t^{k,y}\}_{k \in S_t, y \in Y}$ according to Eq.~(\ref{eq:attensemble}) 
          \For{$j=1, ..., I_g$}
            \State update the generator $\theta$ according to Eq.~(\ref{eq:fedftg-all}) to explore hard samples based on current global model $\omega$
          \EndFor
          \For{$j=1, ..., I_d$}
            \State update the global model $\omega$ according to Eq.~(\ref{eq:fedftg-all}) to transfer the knowledge from $\{\omega_k\}_{k\in S_t}$ to $\omega$
          \EndFor
        \EndFor
    \State return $\omega$, $\theta$
  \end{algorithmic}
\label{alg:fl-server}
\end{algorithm}
\vspace{-0.8cm}
\end{figure}

However, the local models are greatly drifted from each other in data heterogeneity scenario. Thus, traditional gradient averaging could lose the knowledge in local models, and the performance of updated global model is much lower than local models~\cite{zhuang2020performance}. To address this issue, we propose a data-free knowledge distillation method to fine-tune the global model, so that the global model can preserve the knowledge in local models and maintain their performance as much as possible. 
Concretely, the server maintains a conditional generator $G$ that generates pseudo data to capture the data distribution of clients as follows,
\begin{equation}
\small
    \widetilde{x}=G(z,y;\theta),
\end{equation}
where $\theta$ is the parameter of $G$, $z\sim \mathcal{N}(\mathbf{0},\mathbf{1})$ is a standard Gaussian noise, and $y$ is the class label of $\widetilde{x}$ sampled from predefined distribution $p_t(y)$.

As shown in Figure~\ref{fig:overview}, we then input the pseudo data $\widetilde{x}$ to the global model to solve the following problem,
\begin{equation}
\label{eq:l_md}
\small
    \min_{\omega}\mathbb{E}_{z\sim \mathcal{N}(\mathbf{0},\mathbf{1})\atop y\sim p_t(y)} \left [ \mathcal{L}_{md} \right ] = \min_{\omega}\mathbb{E}_{z\sim \mathcal{N}(\mathbf{0},\mathbf{1})\atop y\sim p_t(y)} \left [ \sum_{k\in S_t}\alpha_{t}^{k,y}\mathcal{L}_{md}^k \right ]
\end{equation}
where $\mathcal{L}_{md}^k$ is the model discrepancy between global model $\omega$ and local model $\omega_k$,
\begin{equation}\label{eq:l_md_k}
\small
    \mathcal{L}_{md}^k = D_{KL}(\sigma(D(\widetilde{x};\omega))||\sigma(D(\widetilde{x};\omega_{k}))),
\end{equation}
where $D$ is the classifier. $\sigma$ is the softmax function, which will output the prediction score of $\widetilde{x}$. $D_{KL}$ denotes the Kullback-Leibler divergence. $\alpha_t^{k,y}$ controls the weight of knowledge from different local models during ensemble. By minimizing $\mathcal{L}_{md}$, we transfer the knowledge in local models to the global model. In Section~\ref{sec:3.2} we will introduce how to acquire $p_t(y)$ and $\alpha_t^{k,y}$ to adapt label distribution shift in data heterogeneity scenario. 

\begin{figure}[t]
   \centering
   \includegraphics[width=0.9\linewidth]{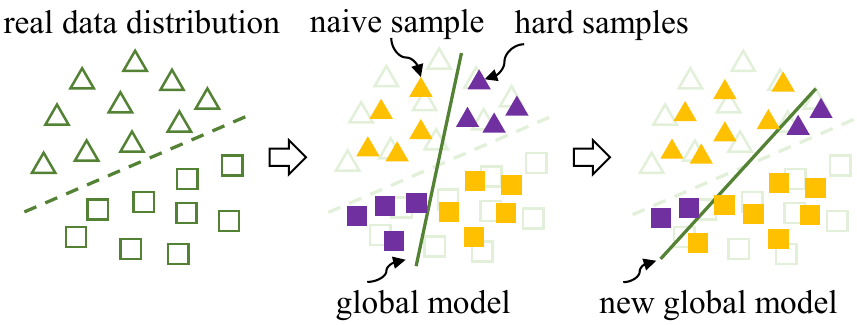}
   \vspace{-0.2cm}
   \caption{Visualization of hard sample mining. By exploring the hard samples in data distribution and fine-tuning the global model, the global model can be gradually corrected during training.}
  \vspace{-0.2cm}
\label{fig:hard_sample_mining}
\end{figure}

\textbf{Data Fidelity and Diversity Constraints.} 
To better extract knowledge from local models, the pseudo data $\widetilde{x}$ should fit the input space of local models. Therefore, we use semantic loss $\mathcal{L}_{cls}$ to train the generator $G$, which will facilitate the fidelity of pseudo data,
\begin{equation}
\label{eq:l_cls}
\small
    \min_{\theta}\mathbb{E}_{z\sim \mathcal{N}(\mathbf{0},\mathbf{1})\atop y\sim p_t(y)} \left [ \mathcal{L}_{cls} \right ] = \min_{\theta}\mathbb{E}_{z\sim \mathcal{N}(\mathbf{0},\mathbf{1})\atop y\sim p_t(y)} \left [ \sum_{k\in S_t}\alpha_{t}^{k,y}\mathcal{L}_{cls}^k \right ]
\end{equation}
where $\mathcal{L}_{cls}^{k}$ is the cross-entropy loss between the prediction of local model on pseudo data $\widetilde{x}$ and the class label $y$,
\begin{equation}\label{eq:l_cls_k}
\small
    \mathcal{L}_{cls}^{k}=\mathcal{L}_{CE}(\sigma(D(\widetilde{x};\omega_{k})),y),
\end{equation}
where $\mathcal{L}_{CE}$ is the cross-entropy loss. By minimizing $\mathcal{L}_{cls}$, $\widetilde{x}$ is enforced to yield higher prediction on class $y$, thus it fits the data distribution of class $y$. 

Simply using $\mathcal{L}_{cls}$ will lead to model collapse of the generator: $G$ outputs the same data for every class. To address this issue, we use diversity loss $\mathcal{L}_{dis}$ in \cite{zhu2021data} to improve the diversity of the generated data,
\begin{equation}\label{eq:l_dis}
\small
    \mathcal{L}_{dis} = e^{\frac{1}{Q*Q}\sum_{i,j\in \{1,...,Q\}}\left (-\left \| \widetilde{x}_i - \widetilde{x}_j \right \|_2 * \left \| z_i - z_j \right \|_2\right )}
\end{equation}
where $\widetilde{x}_i$ is generated using $z_i$. By minimizing $\mathcal{L}_{dis}$, the pseudo data will be diverse and scattered in the data space.

\textbf{Hard Sample Mining.} Training the generator $G$ using $\mathcal{L}_{cls}$ will generate pseudo data $\widetilde{x}$ with low classification error, which means $\widetilde{x}$ contains the most discriminative feature of class $y$ and is easy to be classified. However, these naive samples will not cause the prediction disagreement between global model and local models, i.e., $\mathcal{L}_{md}=0$, thus the global model is not optimized during training. As illustrated in Figure~\ref{fig:hard_sample_mining}, the naive samples are already correctly classified by the global model. To effectively exploit the knowledge in local models and transfer them to the global model, we explore the hard samples in data distribution that cause prediction disagreement between local models and global model. Concretely, we adversarially train the generator and the global model with $\mathcal{L}_{md}$: (1) the generator is enforced to generate hard samples that maximize $\mathcal{L}_{md}$, and (2) the global model is trained to minimize $\mathcal{L}_{md}$ using the hard samples. As a result, the global model can be gradually fine-tuned to fit the data distribution as in Figure~\ref{fig:hard_sample_mining}.

To the end, the overall objective of FedFTG in the server is formulated as an adversarial learning scheme,
\begin{equation}
\small
\!\!\!    \min_{\omega} \max_{\theta} \mathbb{E}_{z\sim \mathcal{N}(\mathbf{0},\mathbf{1}),y\sim p_t(y)} \left [ \mathcal{L}_{md} \!\!-\! \lambda_{cls}\mathcal{L}_{cls} \!-\! \lambda_{dis}\mathcal{L}_{dis} \right ]. \label{eq:fedftg-all}\\
\end{equation}

\subsection{Adaptation to Label Distribution Shift for Effective Knowledge Distillation}\label{sec:3.2}

In data heterogeneity scenario, label distributions are different among clients, i.e., $p^i(y)\neq p^j(y)$ for different clients $i$ and $j$. This indicates that: (1) the local dataset $\mathcal{D}_k$ of client is class-imbalanced, and the local model trained by $\mathcal{D}_k$ contains imbalanced data information; (2) for one class, the importance of knowledge are different among local models of clients. To facilitate more effective knowledge distillation, we propose customized label sampling and class-level ensemble to adapt these two problems respectively. 

 \textbf{Customized Label Sampling.}
Typically, dataset in local client is class-imbalanced in data heterogeneity scenario, even having no data for some classes. It has been proved that deep neural networks tend to learn the majority classes and ignore the minority classes \cite{fang2021layer}.
Hence, the data information of minority classes in local models could be wrong and misleading, and the generated pseudo data are invalid to measure the model discrepancy. If uniformly sample the class label $y$, these invalid data will influence the global model training and induce performance decrease. To mitigate this issue, we customize the sampling probability $p_t(y)$ according to the distribution of whole training data in each round, so that more pseudo data with effective information can be generated,
\begin{equation}
\small
    p_t(y) \propto \sum_{k\in S_t}\sum_{i=1}^{N_k}\mathbb{E}_{(x_k^i, y_k^i)\sim \mathcal{D}_k}\left [ 1_{y_i=y} \right ] = \sum_{k\in S_t}n_k^y,
\label{eq:focusedkd}
\end{equation}
where $1_\mathrm{condition}$ is 1 if the $\mathrm{condition}$ is true and 0 otherwise,  $n_k^y$ is the instance number of class $y$ in client $k$. According to Eq.~(\ref{eq:focusedkd}), the pseudo data of majority classes have high probability to be generated, thus FedFTG can guarantee effective knowledge distillation in data heterogeneity scenario.

\begin{table*}
\caption{Test Accuracy (\%) of different FL methods on CIFAR10 and CIFAR100.}
\small
\vspace{-0.4cm}
\begin{center}\centering
\begin{tabular}{l c c c c c c c c c c}
\hline
 & & \multicolumn{3}{c}{CIFAR10} & &  \multicolumn{3}{c}{CIFAR100} \\
\cline{3-5}\cline{7-9}
& & iid & $\beta=0.6$ & $\beta=0.3$ & & iid & $\beta=0.6$ & $\beta=0.3$ \\
\hline
\textit{centralized learning} & & \multicolumn{3}{c}{92.55$\pm$0.05} & &  \multicolumn{3}{c}{73.98$\pm$0.26} \\
\hline
FedAvg & & 83.78$\pm$0.13 & 82.04$\pm$0.46 & 79.59$\pm$1.01  & & 50.29$\pm$0.34 & 50.67$\pm$0.34 & 50.17$\pm$0.19  \\
FedProx & & 84.10$\pm$0.39 & 82.36$\pm$0.38 & 80.12$\pm$0.43  & & 51.25$\pm$0.62 & 50.94$\pm$0.40 & 50.82$\pm$0.20  \\
FedDyn & & 85.19$\pm$0.58 & 82.87$\pm$0.62 & 80.15$\pm$1.00  & & 53.27$\pm$0.01 & 51.68$\pm$0.31 & 50.51$\pm$0.34  \\
MOON & & 84.34$\pm$0.09 & 82.67$\pm$0.08 & 80.97$\pm$0.46 & & 52.51$\pm$0.70 & 52.55$\pm$0.49 & 51.88$\pm$0.25 \\
SCAFFOLD & & 85.99$\pm$0.06 & 84.55$\pm$0.30 & 82.14$\pm$1.20  & & 53.32$\pm$0.32 & 53.91$\pm$0.33 & 54.36$\pm$0.32  \\
FedGen & & 83.91$\pm$0.36 & 82.23$\pm$0.73 & 79.72$\pm$0.85  & & 50.38$\pm$0.27 & 50.71$\pm$0.55 & 50.08$\pm$0.24  \\
FedDF & & 84.47$\pm$0.20 & 82.92$\pm$0.64 & 80.97$\pm$0.74  & & 52.12$\pm$0.15 & 51.36$\pm$0.02 & 51.26$\pm$0.09  \\
\textbf{FedFTG} & & \textbf{87.34}$\pm$0.16 & \textbf{86.06}$\pm$0.19 & \textbf{84.38}$\pm$0.49  & & \textbf{56.94}$\pm$0.19 & \textbf{56.49}$\pm$0.55 & \textbf{55.96}$\pm$0.39  \\
\hline
\end{tabular}
\end{center}
\label{tab:accuracy}
\vspace{-0.5cm}
\end{table*}

\textbf{Class-Level Ensemble.}
The widely used ensemble method in knowledge distillation assigns the same weight to the knowledge from different teacher models \cite{zhu2021data,lin2020ensemble}, i.e., $\alpha_{t}^{k,y}=\frac{1}{|S_t|}$ in Eq.~\eqref{eq:l_md} and Eq.~\eqref{eq:l_cls}. Due to the label distribution shift, for one class the importance of knowledge are different among local models. If assigning the same weight to clients, the important knowledge can not be figured out and utilized properly. Therefore, we propose class-level ensemble, which assigns the ensemble weight according to the individual data distribution of clients. Specifically, we compute the weight $\alpha_t^{k,y}$ via the data proportion of class $y$ in client $k$ against the total data in $S_t$,
\begin{equation}
\small
    \alpha_t^{k,y} = {n_k^y}\big{/}{\sum_{i\in S_t}n_i^y},
\label{eq:attensemble}
\end{equation}
As a result, the knowledge from clients can be flexibly integrated according to their importance on classes, so that FedFTG can facilitate maximum utilization of knowledge from local models.

\section{Experiments}\label{experiment}
In this section, we empirically verify the effectiveness of FedFTG\footnote{Code is available at https://github.com/ZhangLin-PKU/FedFTG.}. 
We summarize the implementation details in Section \ref{sec:4.1}, and compare FedFTG with several SOTA FL algorithms in Section \ref{sec:4.2}.
Ablation studies are conducted to verify the necessity of each component of FedFTG in Section \ref{sec:4.3}. 
To further validate the effect of FedFTG on real-world FL applications, we evaluate the performance of FedFTG on three real-world datasets in Section \ref{sec:4.4}.

\subsection{Implementation Details}\label{sec:4.1}

\textbf{Baselines.} 
We compare FedFTG against FedAvg~\cite{mcmahan2017communication}, FedProx~\cite{li2018federated}, SCAFFOLD~\cite{karimireddy2020scaffold}, FedDyn~\cite{acar2020federated}, MOON~\cite{li2021model}, FedGen~\cite{zhu2021data} and FedDF~\cite{lin2020ensemble}. Since FedDF does not explain how to obtain the generator, we train it in the same way as FedGen.

\textbf{Datasets.}
CIFAR10 and CIFAR100 datasets \cite{krizhevsky2009learning} with heterogeneous dataset partition are used to test the efficacy of FedFTG, which are two difficult tasks in FL scenario and are widely adopted in FL research. Similar to existing works \cite{acar2020federated,he2020fedml,yurochkin2019bayesian}, we use Dirichlet distribution $\mathbf{Dir}(\beta)$ on label radios to simulate the non-iid data distribution among clients, where a smaller $\beta$ indicates higher data heterogeneity. During the implementation, we set $\beta=0.3$ and $\beta=0.6$. 

\begin{table*}
\caption{Evaluation of different FL methods on CIFAR10 and CIFAR100 ($\beta=0.3$), in terms of the number of communication rounds to reach target test accuracy ($acc$). Note that we highlight the \textbf{best} and \textit{\textbf{second best}} results in bold.}
\small
\vspace{-0.4cm}
\label{tab:round_total}
\begin{center}\centering\small
\begin{tabular}{l c c c c c c}
\hline
\multirow{2}{*}{} & & \multicolumn{2}{c}{CIFAR10} & &  \multicolumn{2}{c}{CIFAR100} \\
\cline{3-4}\cline{6-7}
& & $acc=75\%$ & $acc=80\%$ & & $acc=40\%$ & $acc=50\%$ \\
\hline
FedAvg & & 153.67$\pm$20.33 & 425.33$\pm$61.67  & & 86.67$\pm$6.33 & 713.67$\pm$191.33  \\
FedProx & & 143.67$\pm$0.33 & 391.67$\pm$13.33  & & 86.00$\pm$1.00 & 529.00$\pm$36.00  \\
FedDyn & & \textbf{90.67}$\pm$2.33 & \textbf{183.67}$\pm$23.33  & & 64.00$\pm$8.00 & 239.33$\pm$15.67  \\
MOON & & 128.00$\pm$10.00 & 347.00$\pm$24.00 & & 79.67$\pm$2.33 & 376.00$\pm$29.00 \\
SCAFFOLD & & 100.33$\pm$14.67 & 212.00$\pm$24.00 & & \textbf{\textit{58.33}}$\pm$3.67 & \textbf{\textit{185.67}}$\pm$0.33  \\
FedGen & & 140.00$\pm$4.00 & 406.67$\pm$29.33 & & 95.00$\pm$1.00 & 684.00$\pm$92.00  \\
FedDF & & 132.67$\pm$11.33 & 329.00$\pm$42.00  & & 94.50$\pm$1.50 & 452.00$\pm$5.00  \\
\textbf{FedFTG} & & \textbf{\textit{92.67}}$\pm$14.33 & \textbf{\textit{188.67}}$\pm$31.33  & & \textbf{57.00}$\pm$1.00 & \textbf{166.33}$\pm$10.67  \\
\hline
\end{tabular}
\end{center}
\vspace{-0.4cm}
\end{table*}

\textbf{Network Architecture.}
For both CIFAR10 and CIFAR100, we employ ResNet18 \cite{he2016deep} as the basic backbone. We borrow the generator network architecture from DFAD \cite{fang2019data} for FedFTG and FedDF. For FedGen, the network of generator is composed of two embedding layers (for inputs $z$ and $y$, respectively) and two fully-connected (FC) layers with LeakyReLU and BatchNorm layers between them. 

\textbf{Hyperparameters.}
For all methods, we set the number of local training epoch $E=5$, communication round $T=1000$, the client number $K=100$ with the active fraction $C=0.1$ (i.e., $|S_t|=10$). For local training, the batchsize is $50$ and the weight decay is $1e-3$. The learning rates for classifier and generator are initialized to be $0.1$ and $0.01$ respectively, and they are decayed quadratically with weight $0.998$. 
The dimension of $z$ is $100$ for CIFAR10 and $256$ for CIFAR100. $I$, $I_g$, $I_d$ in Algorithm \ref{alg:fl-server} are $10$, $1$ and $5$, respectively. If not specifically declared, we adopt $\lambda_{cls}=1.0$ and $\lambda_{dis}=1.0$, and adopt SCAFFOLD as the FL optimizer in FedFTG.

We further provide detailed implementations and extra experiment results in the supplementary material.

\subsection{Performance Comparison}
\label{sec:4.2}

\textbf{Test Accuracy.} 
Table~\ref{tab:accuracy} reports the test accuracy of all compared algorithms on CIFAR10 and CIFAR100 datasets.
We provide the performance of centralized learning in the first line.
All experiments are repeated over 3 random seeds. In Table \ref{tab:accuracy}, FedFTG achieves the best performance in all scenarios, surpassing the second one (i.e., SCAFFOLD) by at least $1.5\%$. 
FedDF also employs data-free knowledge distillation to improve the global model in server. It outperforms FedAvg and FedProx, and outperforms FedDyn and MOON in some cases, which further validates the superiority of the scheme ``fine-tuning the global model in server''. However, it is worse than SCAFFOLD and FedFTG.
FedGen yields lower accuracy compared with FedDF and FedFTG, and shows marginal performance gains than FedAvg in some cases. The performance of FedDF and FedGen further verifies the effectiveness of the proposed modules in FedFTG.

\begin{figure}[t]
\vspace{-0.37cm}
\begin{minipage}[b]{0.49\linewidth}
    \centering
    \includegraphics[width=\linewidth]{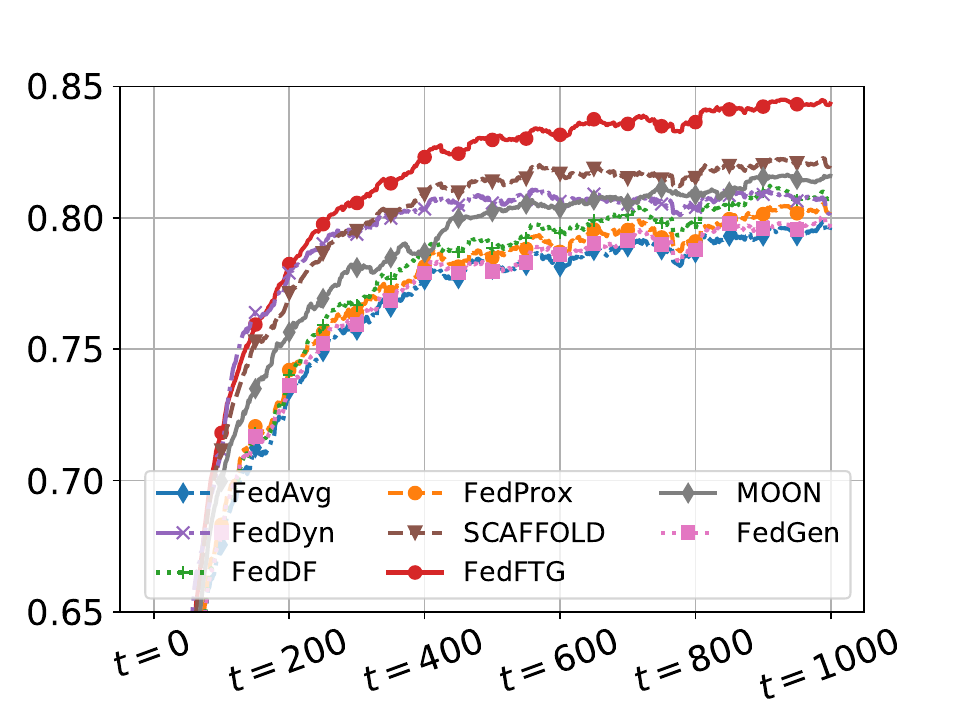}
    \centerline{(a) CIFAR10}\medskip
\end{minipage}
\hfill
\begin{minipage}[b]{0.49\linewidth}
    \centering
    \includegraphics[width=\linewidth]{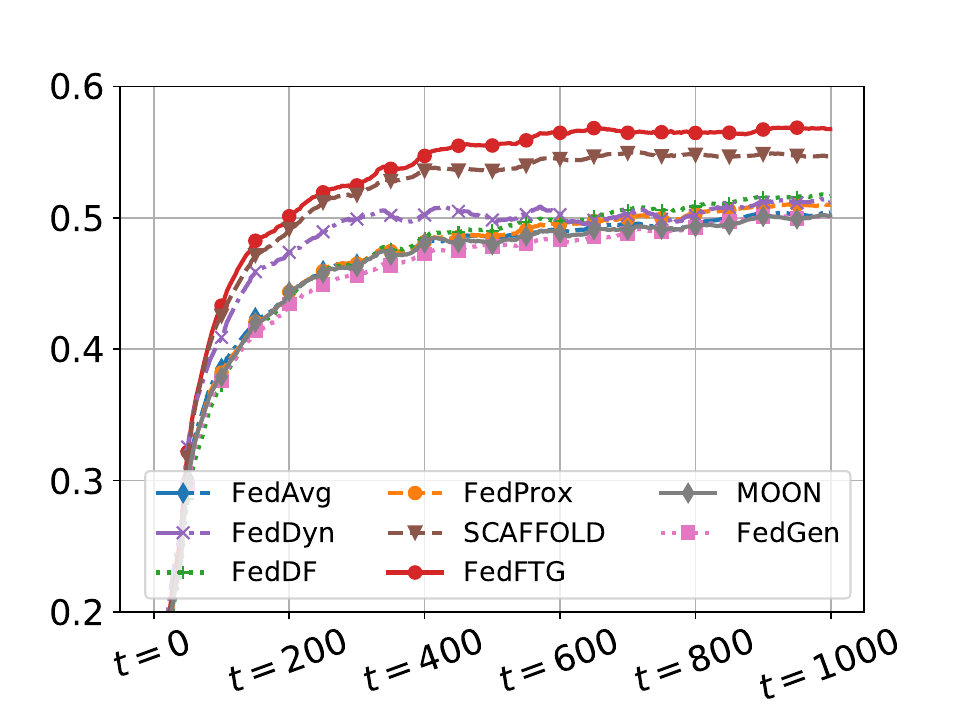}
    \centerline{(b) CIFAR100}\medskip
\end{minipage}
 \vspace{-0.35cm}
\caption{Learning Curve of (a) CIFAR10 and (b) CIFAR100 in 1000 communication rounds ($\beta=0.3$).}
\label{fig:acc_regarding_round}
\vspace{-0.4cm}
\end{figure}

\textbf{Communication Rounds.}
Table \ref{tab:round_total} evaluates different FL methods in term of the number of communication rounds to reach target test accuracy ($75\%$ and $80\%$ for CIFAR10, $40\%$ and $50\%$ for CIFAR100, respectively). In Table \ref{tab:round_total}, FedFTG achieves the second best and the best results on CIFAR10 and CIFAR100, respectively. Besides, FedFTG reduces the round number required by its FL optimizer (SCAFFOLD) in all scenarios. For CIFAR10, although FedDyn uses fewer rounds to achieve the target accuracy, its final accuracy is much worse than FedFTG, as displayed in Table \ref{tab:accuracy}. Below, we provide the results of using FedDyn as the optimizer of FedFTG, and the derived method FedDyn+FedFTG requires fewer rounds to reach target accuracy than FedDyn. Figure \ref{fig:acc_regarding_round} displays the learning curve of different methods in 1000 communication rounds, where FedFTG achieves distinct performance gain after 1000 rounds. Though FedDyn has a faster increase rate in the beginning, the increase trend is gradually slowing down as training goes, and its accuracy is falling behind FedFTG after 150 and 50 rounds for CIFAR10 and CIFAR100 respectively.

\begin{figure}[t]
\vspace{-0.37cm}
\begin{minipage}[b]{0.49\linewidth}
    \centering
    \includegraphics[width=\linewidth]{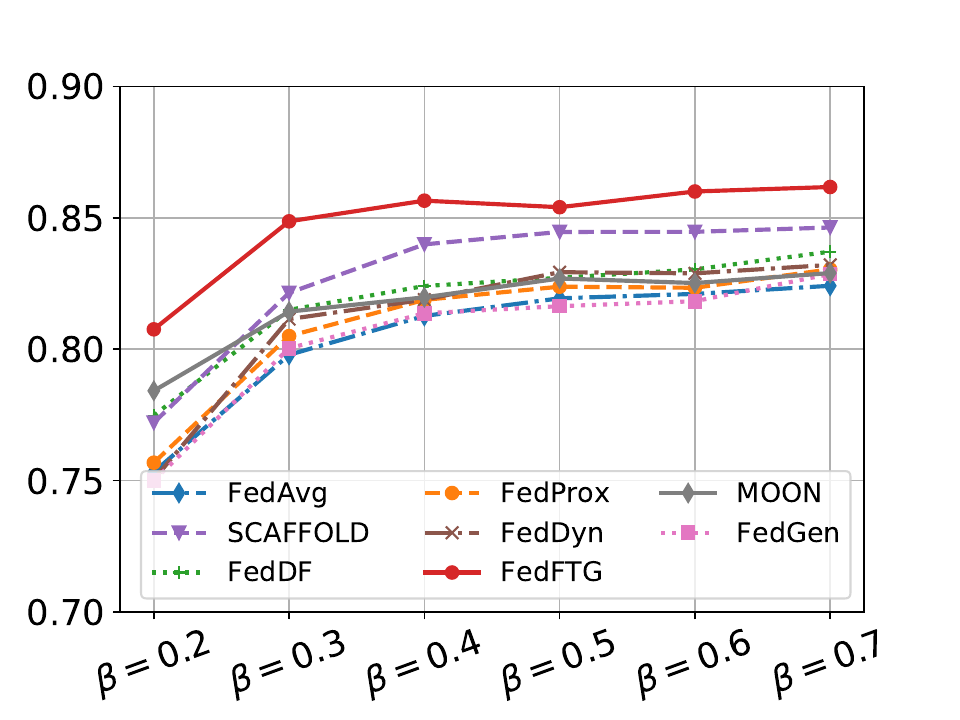}
    \centerline{(a) Test accuracy w.r.t. $\beta$}\medskip
\end{minipage}
\hfill
\begin{minipage}[b]{0.49\linewidth}
    \centering
    \includegraphics[width=\linewidth]{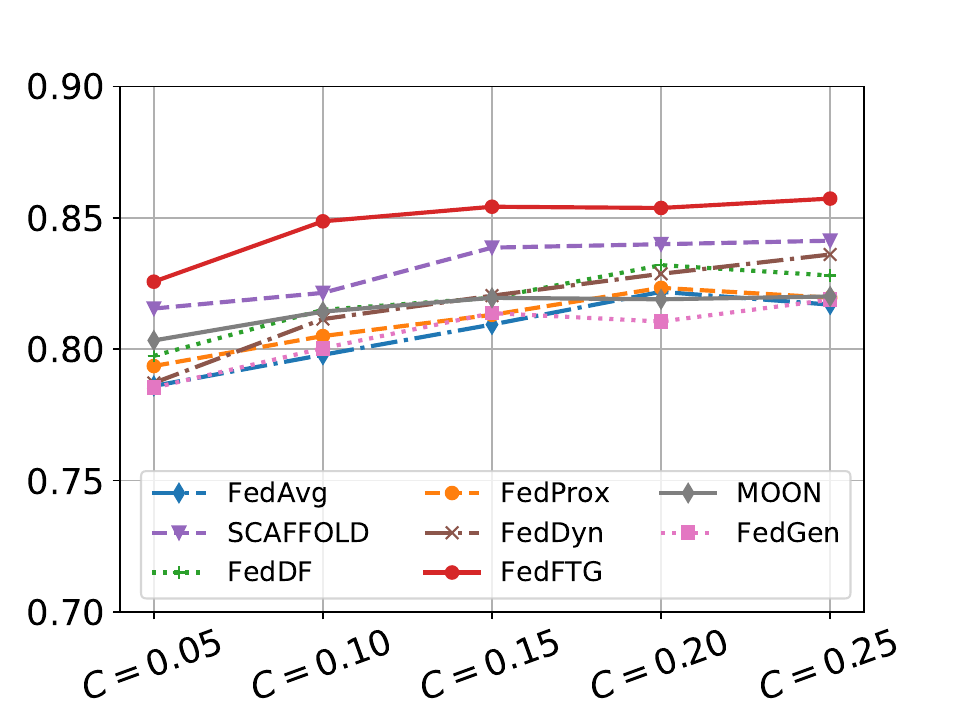}
    \centerline{(b) Test accuracy w.r.t. $C$}\medskip
\end{minipage}
\vspace{-0.35cm}
\caption{(a) Test accuracy w.r.t. data heterogeneity. (b) Test accuracy w.r.t. fraction $C$ of active clients in each round ($\beta=0.3$). All experiments are conducted on CIFAR10.}
\label{fig:acc_regarding_beta_and_frac}
\vspace{-0.47cm}
\end{figure}

\begin{figure*}[t]
\centering
\hfill
\begin{minipage}[b]{0.31\linewidth}
    \centering
    \includegraphics[width=\linewidth]{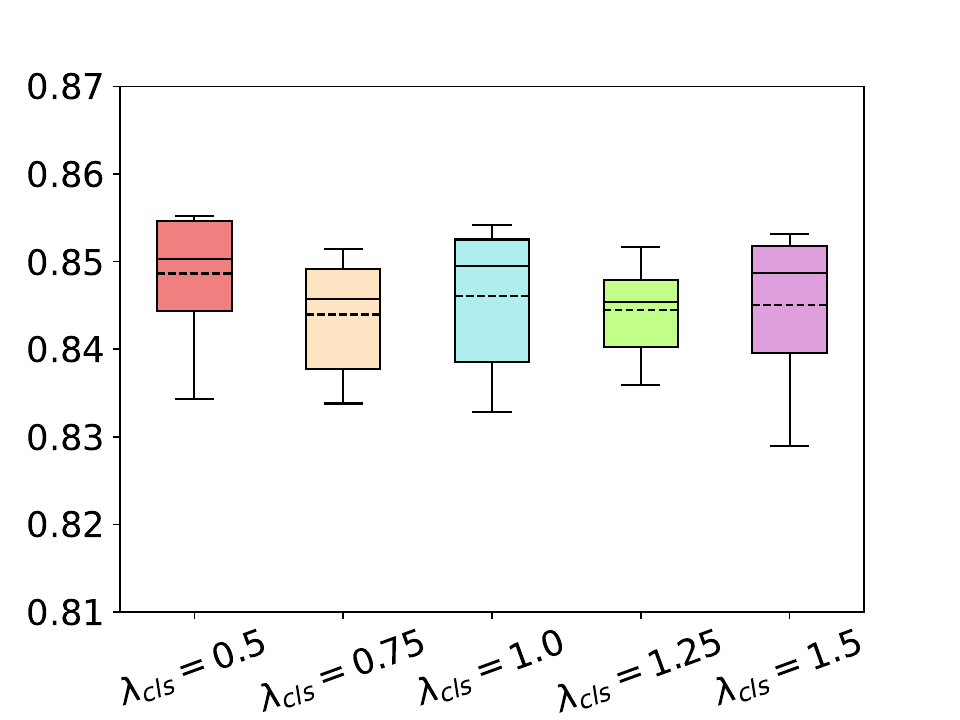}
    \centerline{(a) Box plot w.r.t. $\lambda_{cls}$}\medskip
\end{minipage}
\hfill
\begin{minipage}[b]{0.31\linewidth}
    \centering
    \includegraphics[width=\linewidth]{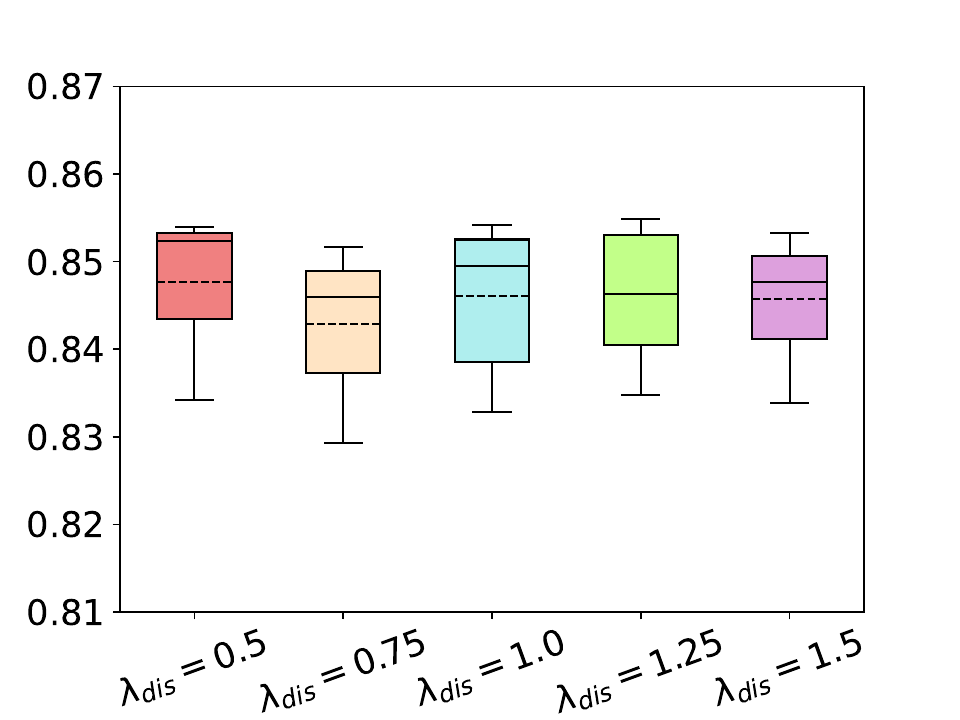}
    \centerline{(b) Box plot w.r.t. $\lambda_{dis}$}\medskip
\end{minipage}
\hfill
\begin{minipage}[b]{0.31\linewidth}
    \centering
    \includegraphics[width=\linewidth]{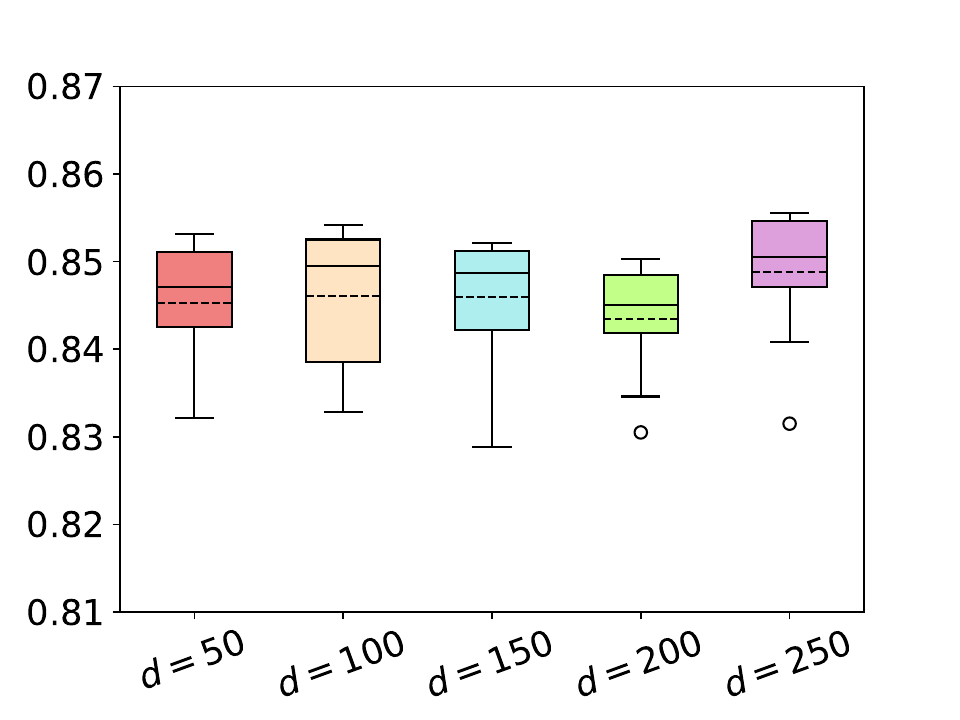}
    \centerline{(b) Box plot w.r.t. $d$}\medskip
\end{minipage}
\hfill
\vspace{-0.3cm}
\caption{Performance of FedFTG using different hyperparameters (a) $\lambda_{cls}$, (b) $\lambda_{dis}$, (c) dimension $d$ of noise $z$ on CIFAR10 with $\beta=0.3$.}
\label{fig:boxplot_regarging_lambda}
 \vspace{-0.4cm}
\end{figure*}

\textbf{Data heterogeneity and Partial Client Participant.}
Figure~\ref{fig:acc_regarding_beta_and_frac}(a) displays the test accuracy on different $\beta$ values. In this figure, FedFTG achieves the best accuracy on all settings, which validates that FedFTG is effective in various data heterogeneity scenarios. Besides, FedFTG gains more accuracy improvement in extreme data heterogeneity scenario $\beta=0.2$. In addition, as the degree of data heterogeneity decreases i.e., $\beta$ increases, the accuracy of each method is ascending. 
Figure \ref{fig:acc_regarding_beta_and_frac}(b) displays the test accuracy of FL methods with different fractions of active clients in each communication round.
FedFTG also yields the best performance in this figure. Besides, the more clients involved in communication, the higher accuracy will be achieved.

\begin{table}[t] 
\caption{The impact of FL optimizer on FedFTG.}
\label{tab:ablation_opitmizer}
\vspace{-0.2cm}
\begin{minipage}[b]{\linewidth}
\begin{center}\centering\small
\centerline{(a) Test accuracy (\%) on CIFAR10, $\beta=0.3$ and $0.6$.}\medskip
\begin{tabular}{l c c c}
\hline
 & & \multicolumn{2}{c}{Accuracy (\%)} \\
\cline{3-4}
& & $\beta=0.6$ & $\beta=0.3$ \\
\hline
FedAvg+FedFTG & & 83.82$\pm$0.31 & 82.27$\pm$0.67 \\
FedProx+FedFTG & & 84.06$\pm$0.32 & 82.21$\pm$0.46 \\
FedDyn+FedFTG & & 83.17$\pm$0.50 & 81.43$\pm$0.18 \\
MOON+FedFTG & & 83.81$\pm$0.45 & 82.19$\pm$0.91 \\
SCAFFOLD+FedFTG & & \textbf{86.06}$\pm$0.19 & \textbf{84.38}$\pm$0.49 \\
\hline
\end{tabular}
\end{center}
\vspace{-0.1cm}
\end{minipage}

\begin{minipage}[b]{\linewidth}
\begin{center}\centering\small
\centerline{(b) the round number to reach the target accuracy}\medskip
\vspace{-0.2cm}
\centerline{($acc=75\%$ and $acc=80\%$) when $\beta=0.3$.}\medskip
\begin{tabular}{l c c c}
\hline
 & &  \multicolumn{2}{c}{Round} \\
\cline{3-4}
& & $acc=75\%$ & $acc=80\%$ \\
\hline
FedAvg+FedFTG & & 122.00$\pm$4.00 & 279.33$\pm$17.67  \\
FedProx+FedFTG & & 117.33$\pm$6.67 & 278.67$\pm$25.33  \\
FedDyn+FedFTG & & \textbf{79.00}$\pm$3.00 & \textbf{168.00}$\pm$13.00  \\
MOON+FedFTG & & 114.33$\pm$6.67 & 276.00$\pm$12.00 \\
SCAFFOLD+FedFTG & & 92.67$\pm$14.33 & 188.67$\pm$31.33 \\
\hline
\end{tabular}
\end{center}
\vspace{-0.6cm}
\end{minipage}
\end{table}

\textbf{Orthogonality of FedFTG with existing FL optimizers.} Table \ref{tab:ablation_opitmizer} provides the performance of FedFTG using FedAvg, FedProx, FedDyn, SCAFFOLD and MOON optimizers. In Table \ref{tab:ablation_opitmizer}, SCAFFOLD+FedFTG yields the best test accuracy among all the optimizers. FedDyn+FedFTG performs better than SCAFFOLD+FedFTG in terms of the round number to reach the target accuracy. This is consistent with the results in Table \ref{tab:round_total}, where FedDyn requires fewer rounds than SCAFFOLD. Comparing Table \ref{tab:ablation_opitmizer} with Tables \ref{tab:accuracy} and \ref{tab:round_total}, we notice that for any FL optimizer, its performance can be largely boosted by using FedFTG. This validates the effectiveness and the orthogonality of FedFTG. Besides, simply using FedAVG+FedFTG as local optimizer already exceeds the other methods in Table~\ref{tab:accuracy} except SCAFFOLD.

\subsection{Ablation Study}
\label{sec:4.3}

\textbf{Necessity of each component in FedFTG.}
Table \ref{tab:ablation_loss} displays the test accuracy of FedFTG after discarding some modules and losses, trained with 500 communication rounds on CIFAR10, $\beta=0.3$. Here \texttt{hsm}, \texttt{cls} and \texttt{abe} represent the hard sampling mining, customized label sampling and class-level ensemble, respectively. We can see that removing any module leads to worse and unstable performance, i.e., lower accuracy and larger confidence interval. In addition, their joint absence can cause a further decrease on accuracy. On the other hand, a similar tendency is observed for the losses: the absence of single loss will lead to performance decrease, and removing multiple losses will enlarge the decrease. 
It should be noticed that, if replacing the KL divergence with Mean Average Square ($\mathcal{L}_{mse}$) to measure the model discrepancy, the model will collapse, which leads to severe performance degradation.

\begin{table}[t]
\begin{center}\centering
\caption{Impact of the each components in FedFTG. The experiments are conducted on CIFAR10, $\beta=0.3$.}
\small
\label{tab:ablation_loss}
\begin{tabular}{l l c}
\hline
\multicolumn{2}{c}{Method} & Accuracy (\%) \\
\hline
baseline & FedFTG & 83.43$\pm$0.10 \\
\hline
\multirow{6}{*}{module} & \texttt{-hsm} & 82.49$\pm$0.42 \\ 
& \texttt{-cls} & 82.39$\pm$0.22 \\
& \texttt{-abe} & 82.40$\pm$0.21 \\
& \texttt{-hsm\&cls} & 82.18$\pm$0.16 \\
& \texttt{-hsm\&abe} & 82.15$\pm$0.14 \\
& \texttt{-cls\&abe} & 82.11$\pm$0.28 \\
& \texttt{-hsm\&cls\&abe} & 81.98$\pm$0.14 \\
\hline
\multirow{4}{*}{loss} & - $\mathcal{L}_{cls}$ & 82.50$\pm$0.55 \\
& \textit{- $\mathcal{L}_{dis}$} & 82.52$\pm$0.35 \\
& \textit{- $\mathcal{L}_{cls}$- $\mathcal{L}_{dis}$} & 82.12$\pm$0.16 \\
& \textit{$D_{KL}\leftarrow\mathcal{L}_{mse}$} & 10.17$\pm$0.26 \\
\hline
\end{tabular}
\end{center}
\vspace{-0.6cm}
\end{table}

\textbf{Robustness of FedFTG on hyperparameters.} To measure the influence of hyperparameter selection, we select $\lambda_{cls}$ and $\lambda_{dis}$ from $\left [ 0.5, 0.75, 1.0, 1.25, 1.5\right ]$ and select the dimension $d$ of noise data $z$ in $\left [ 50, 100, 150, 200, 250\right ]$. Figure \ref{fig:boxplot_regarging_lambda} illustrates the test accuracy in term of the box plot, where FedFTG achieves similar performance among all the choices. Besides, the worst accuracy in Figure \ref{fig:boxplot_regarging_lambda}(a)-(b) is better than the best of previous works in Table \ref{tab:accuracy}. This indicates that FedFTG is not sensitive to the selection of hyperparameter in a large range.

\textbf{Comments on feature-level pseudo data.}
FedGen advises that the data in feature space are more compact than in input space, thus it generates pseudo data in feature-level and fine-tunes the last few FC layers of federated model. Motivated by this, we compare the performance of FedFTG using feature-level generation (\texttt{F}) and input-level generation (\texttt{I}) in Table~\ref{tab:generation_space}. Though FedFTG(\texttt{F}) still exceeds the other methods in Table~\ref{tab:accuracy}, it suffers significant performance drop compared with FedFTG(\texttt{I}), which indicates the input-level generation is more effective for FedFTG. This is because FedFTG(\texttt{F}) only fine-tunes the last few layers of the global model, so the effect of knowledge transfer is limited.

\subsection{Experiments on Real-World Datasets}\label{sec:4.4}

\begin{table}[t] 
\caption{The impact of feature-level generation (\texttt{F}) and input-level generation (\texttt{I}) on FedFTG. The experiments are conducted on CIFAR10, $\beta=0.3$ and $0.6$.}
\label{tab:generation_space}
\vspace{-0.37cm}
\small
\begin{center}\centering
\begin{tabular}{l c c c}
\hline
 & & \multicolumn{2}{c}{Accuracy (\%)} \\
\cline{3-4}
& & $\beta=0.6$ & $\beta=0.3$ \\
\hline
FedFTG(\texttt{F}) & & 84.67$\pm$0.35 & 82.76$\pm$0.81 \\
FedFTG(\texttt{I}) & & 86.06$\pm$0.19 & 84.38$\pm$0.49 \\
\hline
\end{tabular}
\end{center}
\vspace{-0.4cm}
\end{table}

\begin{table}[t]
\caption{Test accuracy (\%) on real-world datasets MIP-TCD, Compcar and Tiny-ImageNet.}
\vspace{-0.37cm}
\small
\label{tab:real_world}
\begin{center}\centering
\begin{tabular}{l c c c}
\hline
Method & MIO-TCD & CompCar & Tiny-ImageNet \\
\hline
FedAvg & 89.63$\pm$1.06 & 43.34$\pm$2.93 & 34.68$\pm$0.67 \\
FedProx & 89.69$\pm$1.00 & 44.07$\pm$3.41 & 35.39$\pm$0.54 \\
FedDyn & 90.47$\pm$0.99 & 50.46$\pm$2.57 & 41.77$\pm$0.28 \\
SCAFFOLD & 89.88$\pm$1.11 & 48.64$\pm$3.46 & 38.80$\pm$0.18 \\
FedGen & 89.85$\pm$1.03 & 45.96$\pm$4.18 & 35.44$\pm$0.35 \\
FedDF & 90.01$\pm$0.70 & 47.31$\pm$3.47 & 36.19$\pm$0.40 \\
\textbf{FedFTG} & {\bf 91.16$\pm$0.92} & {\bf 51.85$\pm$3.46} & {\bf 42.23$\pm$0.22} \\
\hline
\end{tabular}
\end{center}
\vspace{-0.4cm}
\end{table}

In this section, we test the performance of FedFTG on more challenging real-world datasets - vehicle classification datasets MIO-TCD \cite{luo2018mio} and CompCar \cite{yang2015large}, and large-scale image classification dataset Tiny-ImageNet\footnote{https://www.kaggle.com/c/tiny-imagenet}. To better validate the effectiveness of FedFTG, we use the surveillance subset of CompCar, of which the images are collected by surveillance cameras. For MIO-TCD and Tiny-ImageNet, we assign the training data to 100 clients, while for CompCar the client number is 50. $\beta$ of Dirichlet distribution is $0.6$ for all these datasets. The images of MIO-TCD and CompCar are resized to $112*112$ before training, and we adopt a deeper generator for them. The communication round is 50, 100 and 1000 for MIO-TCD, CompCar and Tiny-ImageNet respectively. The other settings are the same as in Section~\ref{sec:4.1}. Experiment results are presented in Table~\ref{tab:real_world}. 

From this table, we find that FedFTG consistently outperforms the other methods in all scenarios, which verifies the effectiveness of FedFTG in real-world FL applications. FedDF and FedGen also adopt data-free knowledge generation to improve the federated model. Though they yield higher performance than FedAvg and FedProx, FedFTG exceeds them by $1\%\sim 6\%$. This further validates the effectiveness of the proposed modules in FedFTG. 
\section{Discussion}

\textbf{Privacy issue.} Since FedFTG recovers the training data of clients in server, it may violate the privacy regulation in FL. However, according to our observation, the pseudo data only captures the high-level feature pattern of real data, which cannot be understood by human beings (see Figure \ref{fig:overview}). Besides, as the generator is trained by all local models, the pseudo data tend to show shared features of data in clients, which means the attribute of individual data will not be revealed. 
Uploading label statistics of data in clients may also leak privacy. One optional solution is adding noise to label statistics. According to our experiments, when the noise ratio is less than 10\%, its influence on performance is less than 0.1\% on CIFAR10, $\beta=0.3$ setting.

\textbf{Communication cost.} Compared with other methods, FedFTG only need to additionally transmit the label statistics of training data (i.e., $\{n_t^{k,y}\}_{k \in S_t, y \in \left [1, .., M \right]}$, $M$ is the class number), which induces negligibly extra transmission cost. 
If the training data keep the same during training, the label statistics can be reported to server before training, thus no extra transmission cost will be introduced.

\textbf{Limitations.} The main limitation of this work mainly exists in computation efficiency. As FedFTG additionally trains the global model apart from local training, it will make the whole training time longer than the other methods. In our experiment, FedFTG requires about double the time of FedAVG in each communication round.
Besides, as the global model training is conducted in the server, FedFTG is more applicable to the cross-silo FL applications as defined in \cite{kairouz2019advances}, where the server can be organizations that own sufficient computation source.

\section{Conclusion}\label{conclusion}

In this paper we propose a new data-free knowledge distillation method FedFTG to fine-tune the global model and to boost the performance of federated learning. 
A hard sample mining scheme is proposed to effectively explore the knowledge in local models and transfer it to global model. Facing the label distribution shift in data heterogeneity scenario, we propose customized label sampling and class-level ensemble to derive maximum utilization of knowledge. Extensive experiments on five benchmarks validate the effectiveness of the proposed FedFTG. 

\vspace{-0.1cm}
\subsubsection*{Acknowledgments}

This work was supported by the National Natural Science Foundation of China under Grant 62088102, and in part by the PKU-NTU Joint Research Institute (JRI) sponsored by a donation from the Ng Teng Fong Charitable Foundation, and in part by Science and Technology Innovation 2030 –“Brain Science and Brain-like Research” Major Project (No. 2021ZD0201402 and No. 2021ZD0201405).

{\small
\bibliographystyle{ieee_fullname}
\bibliography{cvpr2022_zlin}
}

\newpage
\section{Supplementary}

\subsection{Exploration of Long-Tail problem}
\label{sec:appendix-longtail}

\begin{figure}[h]
	\centering
	\vspace{-0.1cm}
	\includegraphics[width=\linewidth,height=3.2cm]{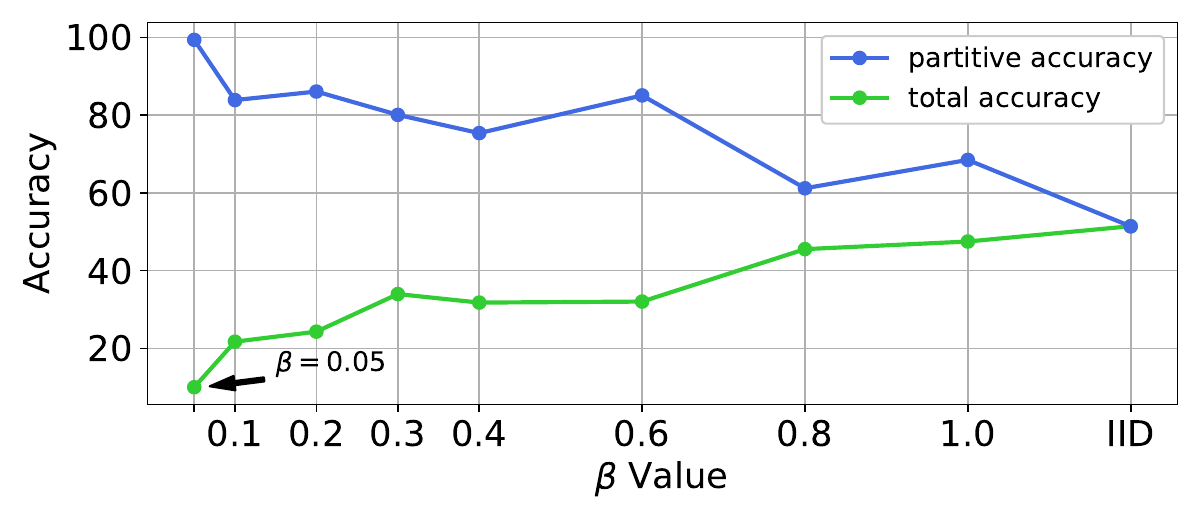}
	\vspace{-0.7cm}
	\caption{Test accuracy of model trained on class-imbalanced data.}
	\label{fig:longtail}
\end{figure}

\begin{figure}[h]
	\centering
	\vspace{-0.3cm}
	\includegraphics[width=\linewidth,height=3.2cm]{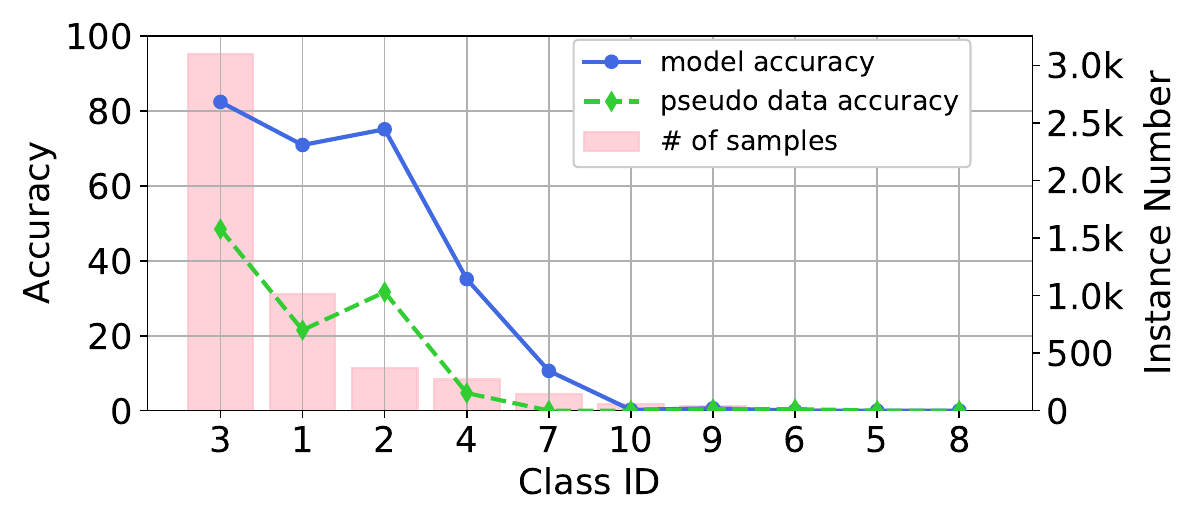}
	\vspace{-0.7cm}
	\caption{Correlation of model accuracy, pseudo data accuracy and the instance number on each class. Note that the class IDs are ordered via the instance number.}
	\label{fig:correlation}
	\vspace{0.2cm}
\end{figure} 

To explore the influence of long-tailed data on model performance, we train models using multiple subsets of CIFAR10, which have different degrees of imbalance. Here the subset is generated by Dirichlet distribution $\mathbf{Dir}(\beta)$, where a smaller $\beta$ indicates more imbalanced data. The data number of each subset is 5000, and the architecture of the model is ResNet34~\cite{he2016deep}. The results are illustrated in Figure~\ref{fig:longtail}. Here, the curves in green and blue are the test accuracy on \textit{total test data} and \textit{partitive test data} respectively, where the distribution of partitive test data is the same as the distribution of training data. We can see there is a performance gap between two curves, and the gap becomes larger when the degree of imbalance is increased. This is because the model only learns the majority classes, and these classes also dominate the partitive test data, thus the model achieves high accuracy on partitive test data; whereas for the total test data that contains balanced data for every class, the model can not correctly predict the data of minority classes, thus the model yields lower test accuracy on total test data. The results in Figure~\ref{fig:longtail} verifies that the model tends to learn majority data from imbalanced training data and ignore the minority classes. In the following, we term the model trained using long-tailed data as ``the biased model''.

To further explore the influence of the biased model on pseudo data generation, we evaluate the accuracy of a biased model trained by a class-imbalanced CIFAR10 subset, and the quality of pseudo data generated via the biased model. The data quality is displayed in terms of the percentage of pseudo data that are correctly classified by a well-trained classifier, which is trained on all data of CIFAR10 and achieves 81.38\% test accuracy. The results are illustrated in Figure~\ref{fig:correlation}. We can see that the model tends to learn majority classes and yields extremely low even zero accuracies for minority classes 7,10 and 9. Moreover, the quality of pseudo data is highly related to original data distribution. For the minority classes, the test accuracy of the pseudo data is less than 10\%, i.e., the quality of the pseudo data is even worse than random noise. This indicates that the pseudo data generated via biased model could be invalid to conduct knowledge transfer, which motivates as to customize the sample probability of label during data generation to facilitate effective knowledge transfer.

\begin{figure*}[h]
\begin{minipage}[b]{0.325\linewidth}
    \centering
    \includegraphics[width=\linewidth]{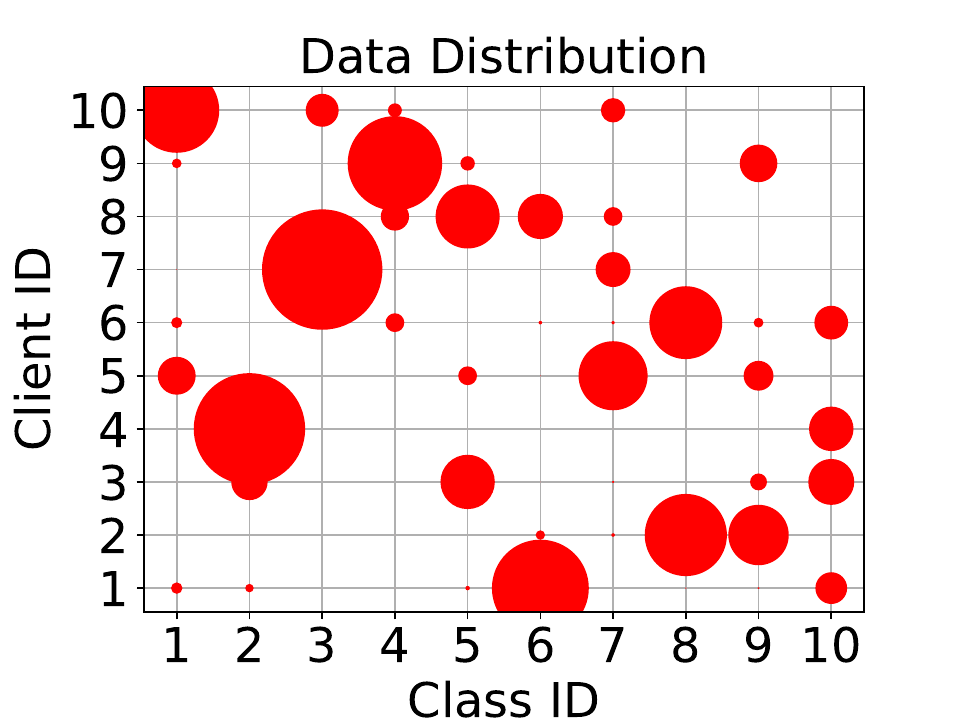}
    \centerline{(a) $\beta=0.1$}\medskip
\end{minipage}
\hfill
\begin{minipage}[b]{0.325\linewidth}
    \centering
    \includegraphics[width=\linewidth]{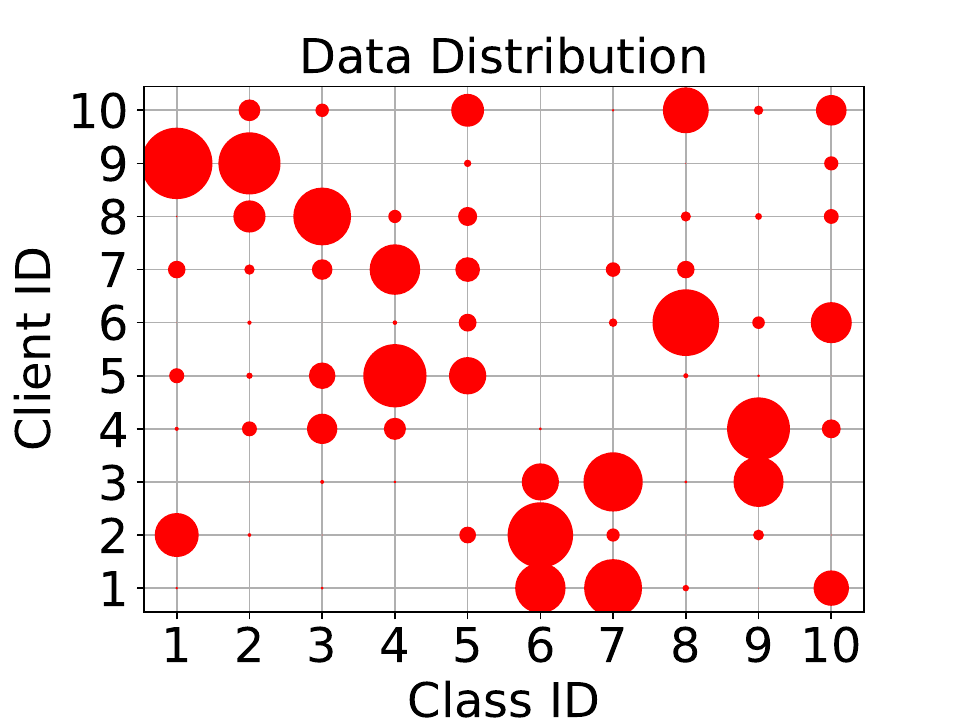}
    \centerline{(b) $\beta=0.3$}\medskip
\end{minipage}
\hfill
\begin{minipage}[b]{0.325\linewidth}
    \centering
    \includegraphics[width=\linewidth]{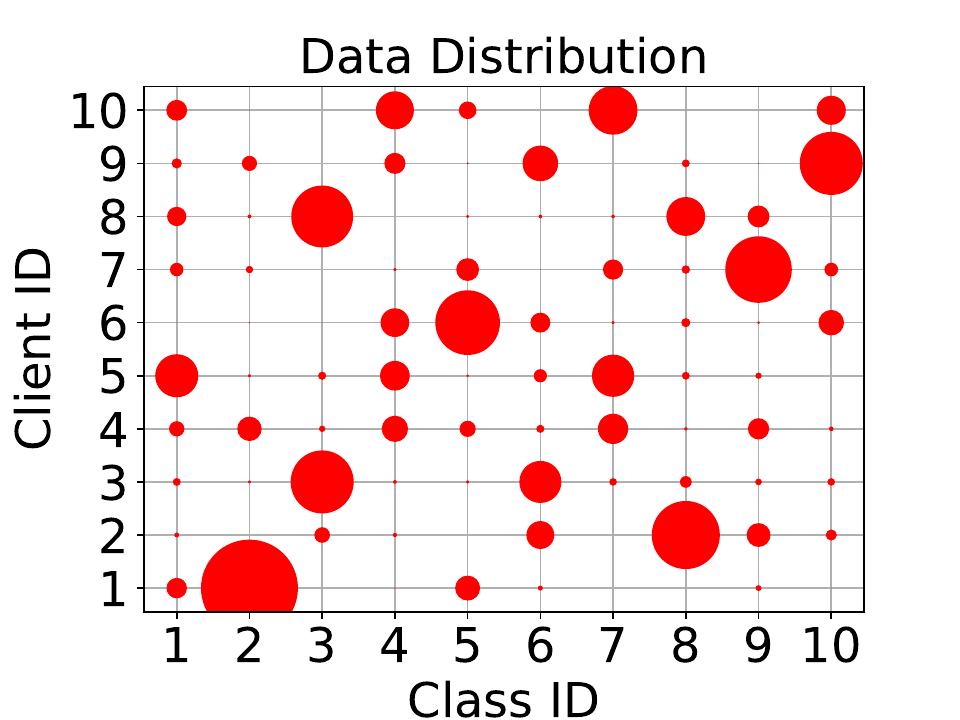}
    \centerline{(c) $\beta=0.6$}\medskip
\end{minipage}
\begin{minipage}[b]{0.325\linewidth}
    \centering
    \includegraphics[width=\linewidth]{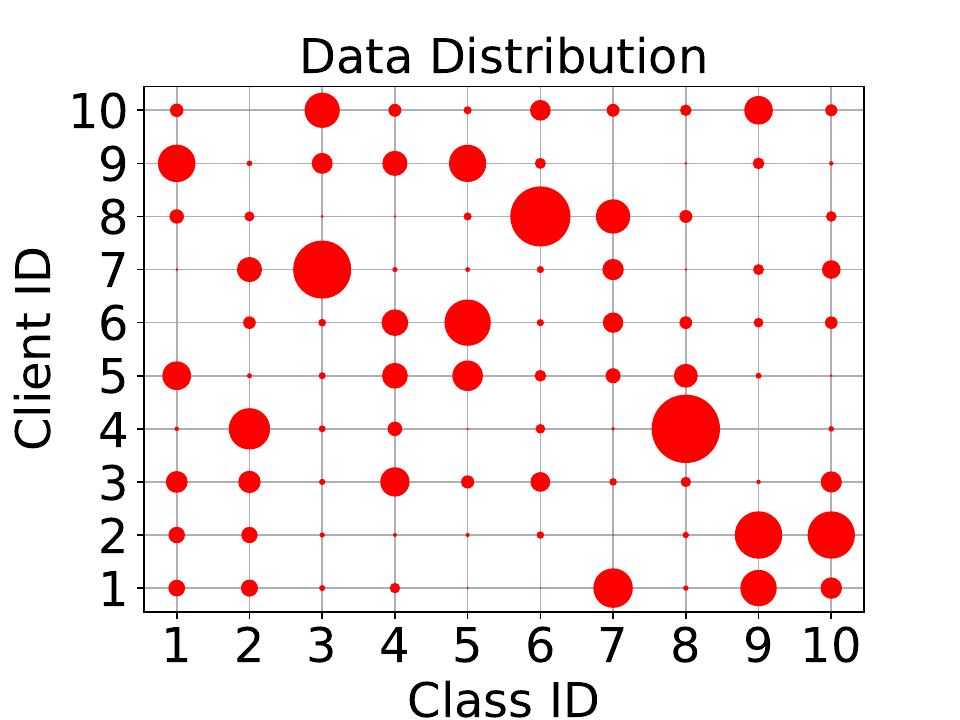}
    \centerline{(d) $\beta=1.0$}\medskip
\end{minipage}
\hfill
\begin{minipage}[b]{0.325\linewidth}
    \centering
    \includegraphics[width=\linewidth]{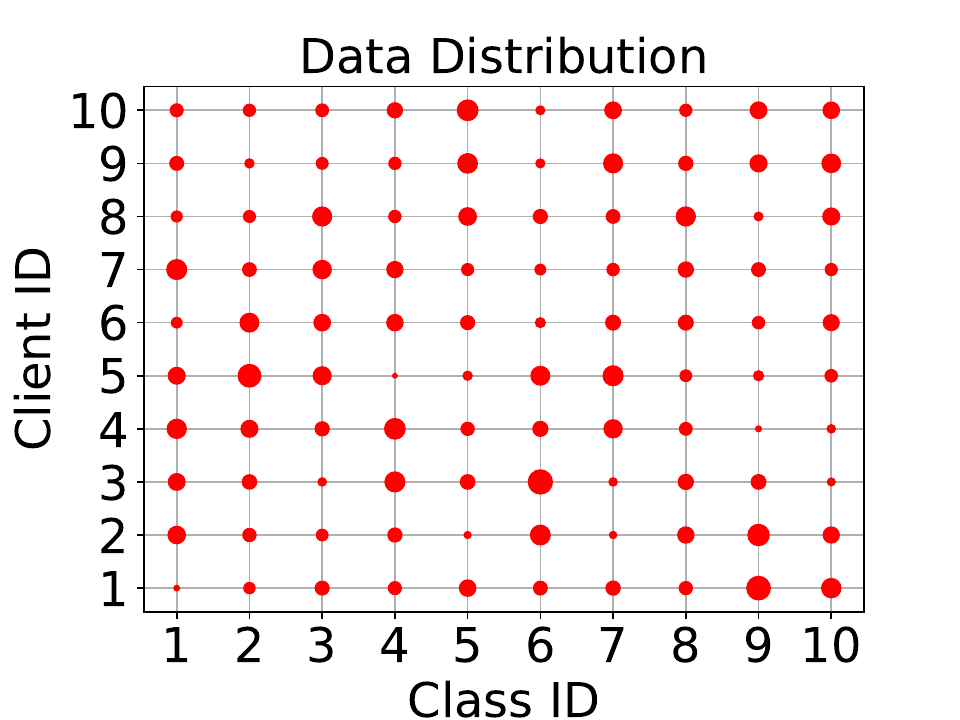}
    \centerline{(e) $\beta=10.0$}\medskip
\end{minipage}
\hfill
\begin{minipage}[b]{0.325\linewidth}
    \centering
    \includegraphics[width=\linewidth]{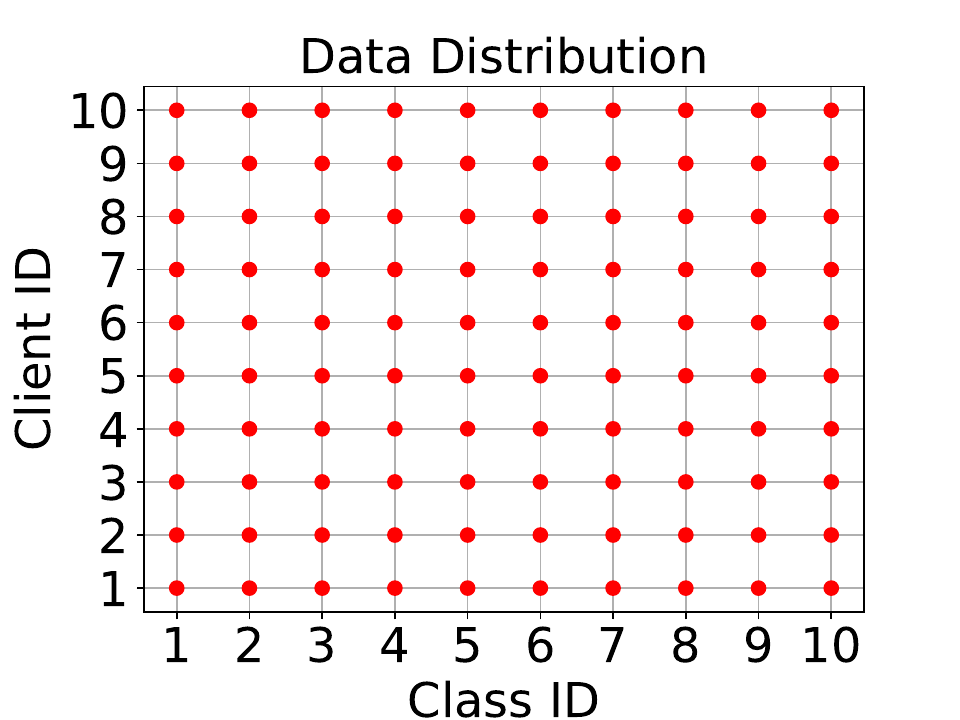}
    \centerline{(f) IID}\medskip
\end{minipage}
\vspace{-0.2cm}
\caption{Visualization of the instance number per class allocated to each clients (indicated by dot size), for different $\beta$ values of Dirichlet distribution $\mathbf{Dir}(\beta)$.}
\label{fig:data_distribution}
\end{figure*}

\begin{table*}[ht]
\caption{The architectures of generators used in Section 4.1 $\sim$ Section 4.3.}
\label{tab:generator_arch}
\begin{minipage}[b]{0.49\linewidth}
\begin{center}\centering\small
\centerline{(a) Generator for FedFTG and FedDF}\medskip
\begin{tabular}{c}
\hline
\vspace{-1pt}$z\in\mathbb{R}^{d}\sim \mathcal{N}(\mathbf{0},\mathbf{1})$ \\
$m=$Map$(y)\in\mathbb{R}^{M},y\in[1,...,M]$ \\
\hline
\vspace{-1pt}FC($z$) $\rightarrow 4096$ \\
FC($m$) $\rightarrow 4096$ \\
\hline
Concat $\rightarrow 8192$ \\
\hline
Reshape, BN $\rightarrow 128\times8\times8$ \\
\hline
Conv2D, BN, LeakyReLU $\rightarrow 128\times8\times8$ \\
\hline
Upsampling $\rightarrow 128\times16\times16$ \\
\hline
Conv2D, BN, LeakyReLU $\rightarrow 64\times16\times16$ \\
\hline
Upsampling $\rightarrow 64\times32\times32$ \\
\hline
Conv2D, Tanh $\rightarrow 3\times32\times32$ \\
\hline
\end{tabular}
\end{center}
\end{minipage}
\hfill
\begin{minipage}[b]{0.49\linewidth}
\begin{center}\centering\small
\centerline{(b) Generator for FedGen}\medskip
\begin{tabular}{c}
\hline
\vspace{-1pt}$z\in\mathbb{R}^{d}\sim \mathcal{N}(\mathbf{0},\mathbf{1})$ \\
$m=$Map$(y)\in\mathbb{R}^{M},y\in[1,...,M]$ \\
\hline
\vspace{-1pt}FC($z$) $\rightarrow 4096$ \\
FC($m$) $\rightarrow 4096$ \\
\hline
Concat, BN $\rightarrow 8192$ \\
\hline
FC, BN, LeakyReLU $\rightarrow 8192$ \\
\hline
FC $\rightarrow 512$ \\
\hline
\\
\end{tabular}
\end{center}
\end{minipage}
\end{table*}

\begin{table*}[ht]
\caption{The architectures of generators used in Section 4.4.}
\label{tab:generator_arch_4.4}
\begin{minipage}[b]{0.49\linewidth}
\begin{center}\centering\small
\centerline{(a) Generator for FedFTG and FedDF}\medskip
\begin{tabular}{c}
\hline
\vspace{-1pt}$z\in\mathbb{R}^{d}\sim \mathcal{N}(\mathbf{0},\mathbf{1})$ \\
$m=$Map$(y)\in\mathbb{R}^{M},y\in[1,...,M]$ \\
\hline
\vspace{-1pt}FC($z$) $\rightarrow s^2$ \\
FC($m$) $\rightarrow s^2$ \\
\hline
Concat $\rightarrow 2s^2$ \\
\hline
Reshape, BN $\rightarrow 512\times \left (s\backslash16 \right )\times \left (s\backslash16 \right )$ \\
\hline
Conv2D, BN, LeakyReLU $\rightarrow 256\times\left (s\backslash8 \right )\times\left (s\backslash8 \right )$ \\
\hline
Conv2D, BN, LeakyReLU $\rightarrow 128\times\left (s\backslash4 \right )\times\left (s\backslash4 \right )$ \\
\hline
Conv2D, BN, LeakyReLU $\rightarrow 64\times\left (s\backslash2 \right )\times\left (s\backslash2 \right )$ \\
\hline
Conv2D, BN, LeakyReLU $\rightarrow 64\times s\times s$ \\
\hline
Conv2D, Tanh $\rightarrow 3 \times s\times s$ \\
\hline
\end{tabular}
\end{center}
\end{minipage}
\hfill
\begin{minipage}[b]{0.49\linewidth}
\begin{center}\centering\small
\centerline{(b) Generator for FedGen}\medskip
\begin{tabular}{c}
\hline
\vspace{-1pt}$z\in\mathbb{R}^{d}\sim \mathcal{N}(\mathbf{0},\mathbf{1})$ \\
$m=$Map$(y)\in\mathbb{R}^{M},y\in[1,...,M]$ \\
\hline
\vspace{-1pt}FC($z$) $\rightarrow s^2$ \\
FC($m$) $\rightarrow s^2$ \\
\hline
Concat, BN $\rightarrow 2s^2$ \\
\hline
FC, BN, LeakyReLU $\rightarrow s^2$ \\
\hline
FC, BN, LeakyReLU $\rightarrow s^2$ \\
\hline
FC $\rightarrow 512$ \\
\hline
\\
\end{tabular}
\end{center}
\end{minipage}
\end{table*}

\subsection{Visualization of Data Heterogeneity}
\label{sec:appendix-heterogeneity}

In Figure~\ref{fig:data_distribution}, we figure out the data distributions of clients that generated by Dirichlet distribution $\mathbf{Dir}(\beta)$ with different $\beta$ as well as IID data distributions. For each $\beta$ value, we display the data distributions of 10 clients. In Figure~\ref{fig:data_distribution}, the data distributions of clients are significantly different when $\beta$ is small, and the client even has no data for some classes. When $\beta$ grows, the data is distributed more evenly in each client, and the discrepancy of data distributions among clients becomes smaller.

\begin{table*}
\caption{Evaluation of different FL methods on CIFAR10 and CIFAR100 ($\beta=0.6$), in terms of the number of communication rounds to reach target test accuracy ($acc$). Note that we highlight the \textbf{best} and \textit{\textbf{second best}} results in bold.}
 \vspace{-0.2cm}
\label{tab:round_total_0.6}
\begin{center}\centering\small
\begin{tabular}{l c c c c c c}
\hline
 & & \multicolumn{2}{c}{CIFAR10} & &  \multicolumn{2}{c}{CIFAR100} \\
\cline{3-4}\cline{6-7}
& & $acc=75\%$ & $acc=80\%$ & & $acc=40\%$ & $acc=50\%$ \\
\hline
FedAvg & & 104.33$\pm$6.67 & 270.67$\pm$13.33  & & 81.67$\pm$2.33 & 563.67$\pm$163.33  \\
FedProx & & 109.67$\pm$8.33 & 263.0$\pm$27.0  & & 81.67$\pm$11.33 & 476.00$\pm$199.00  \\
MOON & & 102.67$\pm$1.33 & 252.33$\pm$32.67 & & 83.67$\pm$3.33 & 354.00$\pm$21.00 \\
FedDyn & & \textbf{72.67}$\pm$7.33 & \textbf{133.33}$\pm$28.67  & & \textbf{\textit{56.00}}$\pm$6.00 & 213.67$\pm$6.33  \\
SCAFFOLD & & 77.00$\pm$3.00 & 161.00$\pm$8.00  & & 61.67$\pm$7.33 & \textbf{\textit{186.33}}$\pm$10.67  \\
FedGen & & 114.00$\pm$8.00 & 284.33$\pm$30.67  & & 82.00$\pm$5.00 & 571.33$\pm$78.67  \\
FedDF & & 97.67$\pm$8.33 & 246.33$\pm$24.67  & & 90.00$\pm$6.00 & 445.00$\pm$42.00  \\
\textbf{FedFTG} & & \textbf{\textit{73.67}}$\pm$4.33 & \textbf{\textit{143.33}}$\pm$5.67  & & \textbf{55.00}$\pm$3.00 & \textbf{152.33}$\pm$10.67  \\
\hline
\end{tabular}
\end{center}
\vspace{-0.27cm}
\end{table*}

\subsection{Detailed Hyperparameters}
\label{sec:appendix-hyperparameter}

Here we introduce the setting of hyperparameters for baselines during experiments. For FedProx, the proximal regularization parameter $\mu$ is $1e-4$. $\alpha$ in FedDyn is $1e-2$. We set the local update round in SCAFFOLD following \cite{acar2020federated}, which is $50$ according to our experiment setting. Following~\cite{li2021model}, we set $\tau=0.5$, tune $\mu$ from $\{0.1, 1, 5\}$ and report the best result. For FedGen and FedDF, the learning rate for the generator is the same as FedFTG, i.e., it is initialized as $0.01$ and is decayed quadratically with weight $0.998$. As Resnet18 only has one fully-connected layer, $l$ in FedGen is $L-1$, where $L$ is the total layer number. 

\subsection{Detailed Architecture of Generator}
\label{sec:appendix-architecture}

Table~\ref{tab:generator_arch} lists the architectures of generators for FedFTG, FedDF and FedGen used in Section 4.1 $\sim$ Section 4.3. Here, $d$ is the dimension of noise data $z$, and it is $100$ and $256$ for CIFAR10 and CIFAR100, respectively. $M$ is the class number of datasets, and it is $10$ and $100$ for CIFAR10 and CIFAR100 respectively. The inplace of LeakReLU is $0.2$ here. Note that in Table~\ref{tab:generator_arch}(b) the output of generator is $512$-dimensional, as the input of the last FC layer in ResNet18 is $512$-dimensional. If using the other classifiers, the dimension of the generator's output should be adjusted accordingly. 

Table~\ref{tab:generator_arch_4.4} lists the architectures of generators used in Section 4.4. Here $d=256$ for all the datasets MIO-TCD, CompCar and Tiny-ImageNet. $s$ is the image size, and $s=112,112,64$ for MIO-TCD, CompCar and Tiny-ImageNet respectively. Note that for the experiments of VGG11 in Table 3 in the main paper, we also adopt these two generators for FedDF, FedGen and FedFTG. 

\subsection{Supplementary Experiment Results}

Table~\ref{tab:round_total_0.6} illustrates the communication rounds of different methods to reach the target test accuracy (75\% and 80\% for CIFAR10, 40\% and 50\% for CIFAR100) when $\beta=0.6$, which is a supplement to Table 2 in the main paper. Same as Table 2, FedFTG achieves the second best and the best convergence for CIFAR10 and CIFAR100 respectively. Besides, it greatly reduces the round numbers required by its FL optimizer SCAFFOLD.

\begin{table}[t]
\small
\caption{Test Accuracy (\%) of different methods on CIFAR10 using VGG11 and ResNet34 networks ($\beta= 0.3$).}
\vspace{-0.4cm}
\begin{center}\centering
\begin{tabular}{l c c c}
\hline
& & VGG11 & ResNet34 \\
\hline
FedAvg & & 82.05$\pm$0.59 & 80.48$\pm$0.89 \\
FedProx & & 82.10$\pm$0.53  & 81.02$\pm$0.53 \\
FedDyn & & 85.38$\pm$0.44 & 81.13$\pm$1.11 \\
MOON & & 83.69$\pm$0.76 & 81.15$\pm$0.46 \\
SCAFFOLD & & 86.78$\pm$0.37 & 83.31$\pm$0.71 \\
FedGen & & 84.38$\pm$0.56
 & 80.72$\pm$0.44 \\
FedDF & & 84.71$\pm$0.78
 & 81.20$\pm$0.46 \\
\textbf{FedFTG} & & \textbf{87.46}$\pm$0.49
 & \textbf{85.00}$\pm$0.45 \\
\hline
\end{tabular}
\end{center}
\label{tab:other_networks}
\vspace{-0.5cm}
\end{table}

Table~\ref{tab:other_networks} displays the test accuracy when adopting VGG11~\cite{simonyan2014very} and ResNet34~\cite{he2016deep} as the classifier. In this table, FedFTG yields the best performance in all scenarios, which validates the effectiveness of FedFTG on various architectures of deep neural network.

\end{document}